\documentclass[a4paper,11pt,final]{article}

\usepackage{epsfig}
\usepackage[cmex10]{amsmath}
\usepackage{amssymb,amsthm}
\usepackage{mathtools}
\usepackage{nccmath}
\usepackage{soul}
\usepackage[normalem]{ulem}
\usepackage{cite}
\usepackage{todonotes}
\usepackage{tikz}
\usepackage{hyperref}
\usepackage{graphicx,amsmath,amssymb,algorithm,algorithmic,textcomp,tikz-cd,float} 
\usepackage{booktabs}
\usetikzlibrary{hobby,decorations.markings,arrows,intersections,shapes}
\usepackage{pgfplots}
\usepgfplotslibrary{fillbetween}
\pgfplotsset{compat=newest}
\usepgfplotslibrary{fillbetween}
\usepackage{verbatim}

\begin{document}

\title{\ \\ \LARGE\bf Information flow and Laplacian dynamics \\on local optima networks}

\author{Hendrik~Richter\textsuperscript{$\diamond$} and Sarah~L.~Thomson\textsuperscript{$\dagger$} \\ \\
\textsuperscript{$\diamond$}\textit{Faculty of
Engineering} \\
\textit{HTWK Leipzig University of Applied Sciences} \\
         Leipzig, Germany  \\    \\ \textsuperscript{$\dagger$} \textit{School of Computing, Engineering and the Built Environment} \\ \textit{Edinburgh Napier University}\\ Edinburgh, United Kingdom
                     \\  \\ Email: 
        \textsuperscript{$\diamond$}hendrik.richter@htwk-leipzig.de ---  \textsuperscript{$\dagger$}s.thomson4@napier.ac.uk. }

\maketitle

\begin{abstract}
We propose a new way of looking at local optima networks (LONs). LONs represent fitness landscapes; the nodes are local optima, and the edges are search transitions between them. Many metrics computed on LONs have been proposed and shown to be linked to metaheuristic search difficulty. These have typically considered LONs as describing static structures. In contrast to this, Laplacian dynamics (LD) is an approach to consider the   information flow across a network as a dynamical process. We adapt and apply LD to the context of LONs. As a testbed, we consider instances from the quadratic assignment problem (QAP) library. Metrics related to LD are proposed and these are compared with existing LON metrics. The results show that certain LD metrics are strong predictors of metaheuristic performance for iterated local search and tabu search. 
\end{abstract}

\section{Introduction}

 Local optima networks (LONs) \cite{ochoa2008study} capture the interplay between an optimisation algorithm and a configuration space. They record the nature of the local optima level in a fitness landscape and have provided insight about search behaviour in combinatorial \cite{verel2010local,ochoa2017understanding,ochoa2014lon} and continuous problems \cite{tomassini2022local,mitchell2023local}. LONs have been analysed from a number of different perspectives. A LON model variant where neutral-fitness connected nodes are compounded into a single node, and where edges between local optima are non-deteriorating (monotonic) in fitness, was introduced and demonstrated on number partitioning \cite{ochoaNPP2017}; these are compressed monotonic LONs: CMLONs for short. The compressed nodes can be viewed as sub-networks in their own right. The CMLON model has been considered in subsequent works on problems such as MAX-SAT \cite{ochoa2020global}, BBOB \cite{mitchell2023local}, feature selection \cite{mostert2019insights}, and parameter tuning \cite{cleghorn2021understanding}. CMLONs have also been constructed for the QAP \cite{thomson2017effect,thomson2023randomness}. Literature has indicated that metrics relating to the nature of these networks can be connected to problem and search difficulty. For example, the original study \cite{ochoaNPP2017} found that the size of compressed nodes decreased with an increase of the number partitioning phase transition parameter; a study on MAX-SAT found the largest compressed node tended to be the global optimum in easier problems. 

Another line of research on LONs is community or cluster detection, where groups of local optima with denser connectivity among themselves than with the rest of the network are identified. This kind of analysis has been conducted on LONs for the QAP \cite{ochoa2011clustering,thomson2017comparing} and NK landscapes \cite{herrmann2016communities}. The existence of multiple clusters --- and the modularity (strength) of them --- has been linked to search difficulty. A related concept is that of \emph{funnels}. In evolutionary computation, a funnel is a fitness landscape structure which can be defined as a basin of attraction at the level of local optima. In the literature, the end of a funnel has been termed a \emph{funnel floor} or a \emph{sink}. These points are attractors and can be optimal or sub-optimal. Measurements related to landscape funnels have gained attention in recent years, perhaps because they can be computed directly from a LON. Funnel metrics, such as the existence of sub-optimal funnels \cite{ochoa2020global}, the incoming flow to the apex of optimal funnels \cite{thomson2023randomness}, or the depth of funnels \cite{thomson2022funnel} seem to have a relationship with problem or search difficulty. Fractal analysis has also been applied to LONs; associated metrics, such as fractal dimension, have been proposed and appear to be linked to metaheuristic performance \cite{thomson2022fractal}.\\
We consider in this article the flow of information encoded within LONs; that is, how the evolutionary dynamics proceed on the fitness landscape. From previous literature, one of the most closely-related works proposed \emph{Markov Chain} local optima networks \cite{chicano2023local}, where transition probability information is added to the model and used to compute hitting times. A full enumeration of the solution space is needed for this approach, which entails scalability issues. There is also the contributions of Herrmann \emph{et al.} on computing pagerank centrality in the context of LONs \cite{herrmann2015predicting,herrmann2018pagerank}. Pagerank is a measurement of centrality typically used for directed graphs; it captures the influence of each node, taking into account directionality of edges. Herrmann's papers demonstrated that the pagerank centrality of the global optimum is highly correlated to search difficulty. \\
In this work, we explore new ways of measuring the flow of information on LONs. A new visualisation method which emphasises the flow between source local optima (at the beginning of search) and sink local optima (search termination points) is presented. Additionally, we look at sampled LONs through a Laplacian lens. To this end, we compute and proposing several metrics associated with Laplacian dynamics (LD). These are shown to have relationships with metaheuristic search through a correlation analysis and algorithm performance prediction models. 

\section{Preliminaries}

 \subsection{Quadratic Assignment Problem}

\paragraph{\bf Definition}

The Quadratic Assignment Problem (QAP)~\cite{koopmans1957assignment} involves assigning $n$ facilities to $n$ locations. The search space is permutations of size $n$; therefore,~$n!$ is the size of this space. In a solution $s$, $s_i$~gives the location of a facility~$i$. There is a distance (cost) between each pair of locations and there is a flow (cost) between each pair of facilities. The costs are specified by the distance matrix $A$ and the flow matrix $B$, which together define an instance.
The objective function associated with a permutation~$s$ is quadratic and considers the sum of pairs of assignment costs (an assignment cost is the product of a distance and a flow cost):%
\begin{equation}%
f(s)=\sum_{i=1}^{n}\sum_{j=1}^{n}{A_{s_{i}s_{j}}B_{ij}}%
\end{equation}%
where $n$~denotes the number of facilities and locations, and where matrix entries are formulated with subscript; for example, \(A_{s_{i}s_{j}}\) is the distance between the locations \(s_i\) and \(s_j\).

\paragraph{\bf Instances}

In this work we use moderate-size (between 25-50 facilities and locations) instances from the well-studied QAP library (QAPLIB)\footnote{\url{http://www.seas.upenn.edu/qaplib/}} \cite{Burkard1997}. From this set, 11 out of 40 instances do not have known global optima (they have not been solved); in these cases, we refer to their best known fitness value as being the global optimum. In general, QAP instances fit into four categories, depending on the nature of their distance and flow matrices \cite{Taillard1995,stutzle2006iterated}: uniform random distances and flows, random flows on grids, real-world problems, and random real-world like problems. The instance group used in the present work contains all four classes.  

\subsection{Monotonic Local Optima Networks} \label{sec:mlon}

\paragraph{\bf Monotonic LON} We describe a monotonic LON (MLON) by a directed graph $G = (L,E)$, where nodes are the local optima $L$, and edges $E$ are the monotonic perturbation edges. 

\paragraph{\bf Local optima}  We assume a search space $S$ with a fitness function $f$ and a neighbourhood function $N$; these comprise the fitness landscape. A local optimum $l \in L$, which in the QAP is a minimum, is a solution $l$ such that $\forall s \in N(l)$, $f(l) \leq f(s)$. Notice that the inequality is not strict --- this accounts for the possible presence neutrality (local optima of equal fitness).

\paragraph{\bf Monotonic perturbation edges} Edges $E$ are directed and based on the perturbation operator ($k$-exchange, $k > 2$). There is an edge from local optimum $l_1$ to local optimum~$l_2$, if $l_2$ can be obtained after applying a random perturbation ($k$-exchange) to $l_1$ followed by local search, and $f(l_2) \leq f(l_1)$. These edges are called {\em monotonic} as they record only non-deteriorating transitions between local optima. Edges are weighted with estimated frequencies of transition. The weight is the number of times a transition between two local optima basins occurred with a given perturbation.
In this work, we consider a set of monotonic local optima networks from previous literature \cite{thomson2022fractal}. For problems of any realistic size, a full enumeration of local optima and edges between them is not possible. These are therefore \emph{sampled} LONs. Further details of their construction will be provided in Section \ref{sec:LONconstr}.

\section{Methodology}
\subsection{Laplacian dynamics on directed graphs}

As defined in Sec. \ref{sec:mlon}, LONs are described by a directed graph (digraph) $G=(L,E)$ with nodes $\ell_i \in L$, $i=1,2,\ldots,n$, representing local optima $l_i$ and edges $e_{ij} \in E$ describing perturbation edges. The edges are directed so that $e_{ij}$
  implies an edge from $\ell_i$ to $\ell_j$. We may interpret the graph as a model of how information flows over the edges from node to node. Thus, we may informally say that 
an directed edge  $e_{ij}$ means information goes from node $\ell_i$ to node $\ell_j$ ($\ell_i$ influences $\ell_j$), but also that $\ell_j$ gets information from $\ell_i$  ($\ell_j$ reacts on, or is influenced by, or ``sees'', $\ell_i$).

For the following discussion, some definitions about structure and connectedness in directed graphs are needed. 
Typically, LONs are weakly connected graphs. It means the underlying graph (obtained by ignoring the direction of edges) is connected. Weakly connected implies there are no isolated nodes (isolated nodes may have self-loops, but otherwise have neither incoming nor outgoing edges connecting them to other nodes), but also that there are nodes which are not reachable from any other node of the LON~\cite{bang09}. Moreover, a LON may have strictly connected components (SCCs). A SCC is a subgraph of the LON where for  every ordered pair of nodes, $\ell_i$ and $\ell_j$, there is a directed path not only from $\ell_i$  to $\ell_j$ but also from $\ell_j$  to $\ell_i$.  Thus, every node of a SCC is reachable from any other node of the SCC, which implies that for monotonic LONs all nodes in a SCC have the same fitness. We notice that SCCs are related to compressed LONs (CMLONs) \cite{ochoaNPP2017}. While SCCs consider direction in their criteria, the CMLON model allows a plateau of local optima to be compressed into one node if they form a \emph{weakly} connected component; that is, if there is a path between the nodes --- regardless of direction. 

LONs describe how evolutionary dynamics on fitness landscapes moves towards better fitness. This implies source and sink nodes~\cite{ochoaNPP2017,thomson2017comparing}.
A source node is a node with no incoming edge from another node (a self-loop is permitted). Source nodes are the initial optima and thus constitute the starting points of evolutionary search.  A sink node is a node with no outgoing edge to another node, again a self-loop is possible. Sink nodes represent optima where no further improvement has been found and thus the terminal points of evolutionary search.  
Source and sink nodes may also group as SCCs. A source SCC is a SCC with no ingoing edge from a node outside the SCC. A sink SCC is a SCC with no outgoing edge to a node outside the SCC.

Dynamical processes on $G$ can be described by Laplacian operators~\cite{newman10}. For defining these operators we need the (combinatorial) adjacency matrix $A$ with elements $a_{ij}>0$ indicating an edge $e_{ji}$ from $\ell_j$ to $\ell_i$. If a node $\ell_i$ has no incoming edges, we impose a loop by $a_{ii}=1$. Thus, the in-degree  $d_i=\sum_{j=1}^{n} a_{ij}$ is non-zero for all nodes, 
we have a  non-singular (and consequently invertible) in-degree matrix $D=\rm{diag}$ $(d_1,d_2,\ldots,d_n)$, and can define the random walk Laplacian by\begin{equation} 
\mathcal{L}=I-D^{-1}A.
\end{equation}
Based on the random walk Laplacian $\mathcal{L}$, we have two dynamical process on $G$ (which are dual to each other) by the first order Laplacian differential equations 
\begin{align}
\dot{p}&=-\mathcal{L}p  \label{eq:cons}\\
\dot{q}&=-q\mathcal{L}. \label{eq:diff}
\end{align}
The process \eqref{eq:cons} is frequently called consensus, while the process \eqref{eq:diff} is known as diffusion. 
 For understanding the dynamics on the directed graph $G$, we are interested in the solutions of the Laplacian differential equations \eqref{eq:cons} and \eqref{eq:diff}, which are $p(t)= \exp{(-\mathcal{L}p)}p_0$ and $q(t)=q_0\exp{(-q\mathcal{L})}$, and particularly in the asymptotic dynamics $\lim \limits_{t \to \infty} p(t)$ and $\lim \limits_{t \to \infty} q(t)$. Both processes describe how information flows on the graph, but with an opposite directional focus. 
In other words, the dual processes of consensus and diffusion describe 
for the whole graph where information comes from and goes to. The consensus problem specifies how information flows according to the orientation of the graph and spreads over the entire graph. It eventually tracks the sinks of information. Diffusion, on the other hand, characterizes  the flow contrary to the orientation of the graph, and thus reversed to the flow of information. It thus tracks the sources of information.

\begin{figure}[htb]
\begin{center}
\includegraphics[trim = 55mm 110mm 50mm 110mm,clip, width=6.2cm, height=3.5cm]{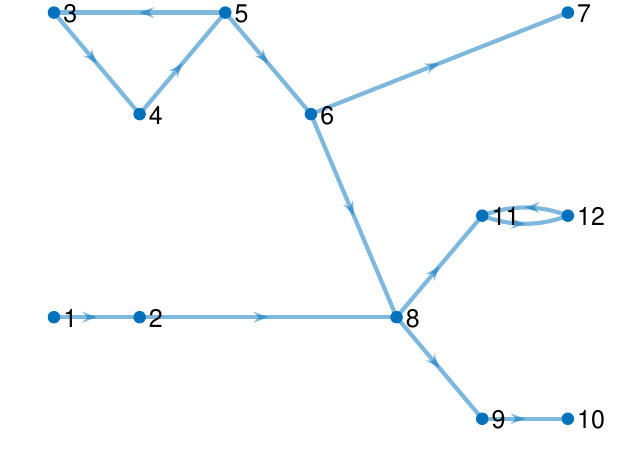}

\caption{Simple example of a LON for illustrating LD}
\label{fig:LON_graph}
\end{center}
\end{figure}

Recently, it was shown by Veerman \& Kummel~\cite{veer19} that for weakly connected digraphs the asymptotic solutions of the Laplacian differential equations \eqref{eq:cons} and \eqref{eq:diff} can be obtained by the left and right kernels of the Laplacian $\mathcal{L}$. We write $\Gamma_{left}$ for the left kernel and $\Gamma_{right}$  for the right kernel with $\Gamma_{left} \cdot \mathcal{L}=\mathbf{0}^T$ and $\mathcal{L} \cdot \Gamma_{right}=\mathbf{0}$. The asymptotic solutions are $\lim \limits_{t \to \infty} p(t)=\Gamma p_0$ and  $\lim \limits_{t \to \infty} q(t)=q_0 \Gamma$ with $\Gamma=\Gamma_{right} \otimes \Gamma_{left}$. The solutions  imply that for each node $\ell_i$ we can  track sinks and sources of information on the LON by setting initial conditions $p_0$ and $q_0$ and analyzing $\Gamma p_0$ and $q_0 \Gamma$. As a simple example consider the LON in Fig. \ref{fig:LON_graph}. The LON has 4 source nodes ($\ell_1$,$\ell_3$-$\ell_5$) and 4 sink nodes ($\ell_7$,$\ell_{10}$-$\ell_{12}$). Three of the source nodes ($\ell_3$-$\ell_5$) form a strictly connected component (SCC), which also applies to two of the sink nodes ($\ell_{11}$-$\ell_{12}$). We consider the LON with unity weights, that is for all $a_{ij}>0$, we have $a_{ij}=1$. With the matrix 
\begin{equation*}
\Gamma= \scriptsize{\frac{1}{6}
\left(\begin{array}{cccccccccccc} 6 & 0 & 0 & 0 & 0 & 0 & 0 & 0 & 0 & 0 & 0 & 0\\ 6 & 0 & 0 & 0 & 0 & 0 & 0 & 0 & 0 & 0 & 0 & 0\\ 0 & 0 & 2 & 2 & 2 & 0 & 0 & 0 & 0 & 0 & 0 & 0\\ 0 & 0 & 2 & 2 & 2 & 0 & 0 & 0 & 0 & 0 & 0 & 0\\ 0 & 0 & 2 & 2 & 2 & 0 & 0 & 0 & 0 & 0 & 0 & 0\\ 0 & 0 & 2 & 2 & 2 & 0 & 0 & 0 & 0 & 0 & 0 & 0\\ 0 & 0 & 2 & 2 & 2 & 0 & 0 & 0 & 0 & 0 & 0 & 0\\ 3 & 0 & 1 & 1 & 1 & 0 & 0 & 0 & 0 & 0 & 0 & 0\\ 3 & 0 & 1 & 1 & 1 & 0 & 0 & 0 & 0 & 0 & 0 & 0\\ 3 & 0 & 1 & 1 & 1 & 0 & 0 & 0 & 0 & 0 & 0 & 0\\ 3 & 0 & 1 & 1 & 1 & 0 & 0 & 0 & 0 & 0 & 0 & 0\\ 3 & 0 & 1 & 1 & 1 & 0 & 0 & 0 & 0 & 0 & 0 & 0 \end{array}\right)}
\end{equation*}
specifying the solution of the Laplacian differential equations, we can track the information flow on the LON in Fig. \ref{fig:LON_graph} as follows.

For analyzing the information flow either from or to a node $\ell_i$, we set the initial condition $\mathbf{1}(\ell_i)$, which concentrates the effect on this node and is an unit vector with a unity element at position $i$ and zero elements otherwise. For instance, for the local optimum associated with the first node $\ell_1$, we have  $\mathbf{1}(\ell_1)=(1 \: 0 \:0 \:0 \:0 \:0 \: 0 \: 0 \: 0 \: 0 \: 0 \: 0)^T$. Thus, according to consensus the quantity $\Gamma \mathbf{1}(\ell_1)=\small{\frac{1}{6}}(6 \: 6 \:0 \:0 \:0 \:0 \: 0 \: 3 \: 3 \: 3 \: 3 \: 3)^T$ describes the information flow from $\ell_1$ to the entire LON. We see that $\ell_1$ influences itself and the nodes $\ell_2$ as well as $\ell_8$--$\ell_{12}$ which are all the nodes that are reachable by directed paths connected to $\ell_1$. Only from source nodes (which can be single nodes or nodes in SSCs) information flows to the LON. Thus, we have a metric on the information flow for each source node by the corresponding columns of $\Gamma$.   Furthermore, we get an overall metric of the influence that each source node has on the graph by averaging the information flow for the node \cite{mas10}. Thus,  we take the column-average of $\Gamma$, which is known as the influence vector 
\begin{equation} I=\frac{1}{n} \sum_{i=1}^n \Gamma \mathbf{1}(\ell_i). \label{eq:influe} \end{equation}
For the LON in Fig. \ref{fig:LON_graph}, we have $I=\small{\frac{1}{24}}(9 \: 0 \:5 \:5 \:5 \:0 \: 0 \: 0 \: 0 \: 0 \: 0 \: 0)$. In this view, the local optima in node $\ell_1$ has almost twice as much outgoing information flow, and thus influence on the LON, as the optima in $\ell_3$-$\ell_5$. Any other node only conveys information coming from source nodes and thus has no influence of its own. 

For tracking the sources of information on the LON we analyze diffusion. In principle, we can track information for every node in the LON, but most interesting are the sink nodes, which correspond with best solutions found on the underlying fitness landscape.  For instance, for the sink node $\ell_{12}$ we obtain $ \mathbf{1}(\ell_{12})\Gamma=\small{\frac{1}{6}}(3 \: 0 \:1 \:1 \:1 \:0 \: 0 \: 0 \: 0 \: 0 \: 0 \: 0)$. In other words, information flowing to $\ell_{12}$ origins from $\ell_1$ as well as from $\ell_3$-$\ell_5$, or put differently, the influencers of $\ell_{12}$ are $\ell_1$ and $\ell_3$-$\ell_5$.   By analyzing the rows of $\Gamma$, we see that the same applies to $\ell_{10}$, but the influencers of $\ell_7$ are just $\ell_3$-$\ell_5$. Another interesting result about diffusion is that it preserves probability~\cite{veer19}. Thus, $\mathbf{1}(\ell_i) \Gamma$ gives a probabilistic measure of the information flow ending in $\ell_i$. For $\ell_{12}$ this means it is 3 times more likely that the information flow comes from $\ell_1$ than from either  $\ell_3$,  $\ell_4$, or  $\ell_5$. However, as the probabilities are preserved  (the row sums  equal unity), $\mathbf{1}(\ell_i) \Gamma$  only accounts for the relative influence which a specific node $\ell_i$ receives, but not for comparing the effects over the nodes of the LON.  

\subsection{LON metrics from Laplacian dynamics}\label{laplacian_metrics_description}

In the following we relate Laplacian dynamics (LD) to LONs.
As LONs are tools for studying evolutionary dynamics, differences in the information flow on LONs may be helpful for explaining differences in the search dynamics of the underlying fitness landscapes. For instance, we would expect
more easily searchable fitness landscapes to have LONs where the influence on sink nodes with better fitness is higher than on sink nodes with poorer fitness. Moreover, the edges of the LONs we consider are perturbation edges. According to the iterative local search used with the QAP, see Sec.~\ref{sec:LONconstr}, an initial solution is generated by a random assignment of items to locations. Thus, we can assume that finding an initial optimum (a source node or node in a source SCC) occurs with the same probability over all source nodes. In other words, the flow starts with the same probability in any source node. Thus, a more easily searchable fitness landscape would be characterized by source node which have a
rather balanced influence on the LON and not with large differences in the elements of the influence vector.   In the following, we attempt to formalize these properties by metrics based on Laplacian kernels and particularly on the matrix $\Gamma$.

Our first metrics are consensus-based, directly use the influence vector~\eqref{eq:influe} and basically capture differences of influence over source nodes. Therefore, we calculate the variance of the elements of the influence vector. We use two variants. The first, $\mathcal{I}_1$,  considers the whole influence vector and thus not only counts the influence of the source nodes, but takes into account the zero elements of all other nodes as well. Thus, in some ways the metric $\mathcal{I}_1$ also reflects the ratio between the number of source nodes and the total number of nodes in the LON. The second variant,  $\mathcal{I}_2$, ignores the zero elements of the influence vector by only considering the source nodes. Thus,   
we have 
\begin{align}
\mathcal{I}_1&=Var(I), \label{eq:i1} \\
\mathcal{I}_2&=Var(I>0).
\end{align}
A second metric is diffusion-based and focuses on the influence of the network on sink nodes, which can include the global optima of the underlying optimization problem. Therefore, we define the reduced influence vector $I_{red}$, which uses the column-average of $\Gamma$ for the sink nodes only. With the number of sink nodes $n_{sink}$, we get the metric $\mathcal{I}_3$  measuring the difference between the reduced influence vector and the influence vector: 
\begin{align} I_{red}&=\frac{1}{n_{sink}} \sum_{i=1}^{n_{sink}}  \mathbf{1}(\ell_i) \Gamma, \label{eq:influe_red} \\
 \mathcal{I}_3&=\|I_{red}-I\|. \end{align}
The next set of metrics explicitly involves the fitness $f$. 
As the QAP is a minimization problem, small values of the fitness $f_i$ are superior. In order to consider fitness of the nodes in metrics related to the influence vector, it is numerically convenient to normalize fitness to the interval $[0,1]$ and have a reversal of fitness values. Thus, we define  
$F_i$, $i=1,2,\ldots,n$, as reversed normalized fitness. Thus, we get $F_i=1$ for the node with the smallest fitness $\displaystyle\min_{i} f_i$, $F_i=0$ for the node with the largest fitness $\displaystyle\max_{i} f_i$ and $0<F_i<1$ for all other nodes. With the reversed normalized fitness we calculate  a metric $ \mathcal{I}_4$,  which is  the difference between the fitness-weighted reduced influence vector and the influence vector:
\begin{align} I_{red}^f&=\frac{1}{n_{sink}} \sum_{i=1}^{n_{sink}} F_i \cdot  \mathbf{1}(\ell_i) \Gamma, \label{eq:influe_red_f} \\
 \mathcal{I}_4&=\|I_{red}^f-I\|.
\end{align}
Finally, we weight the influence vector by the fitness values to get the measure 
\begin{equation} \mathcal{I}_5=I \times F \label{eq:i5}\end{equation}
which accounts for the relation between influence and fitness of the source nodes.

So far the metrics are based on the influence vector accounting for the influence of source nodes on the graph. In other words, the metrics the metrics 
$\mathcal{I}_1$--$\mathcal{I}_5$ can be viewed as source-oriented metrics.  As the flow on the LON is directed towards the sink nodes, it appears desirable to have an alternative perspective: sink-oriented metrics. This seems to be particularly natural to the approach we are proposing as the Laplacian processes described by the differential equations \eqref{eq:cons} and \eqref{eq:diff} explicitly offer such a dual view.

In this line of thinking, we next analyse the 
reverse graph (or transpose graph) $G^R$ of the graph $G=(L,E)$ describing the LON. 
The reverse LON $G^R$ contains the same node set as the LON $G$, but all edges $e_{ij}$ have reversed directions $e_{ji}$. Thus, while a LON connects source nodes to sink nodes,  the reverse graph of the LON connects the sink nodes to the source nodes. The analysis with Laplacian operators as described above can be carried out accordingly and we get a matrix $\Gamma^R$ for specifying the solutions of the reverse Laplacian differential equations. Thus, analogously to the influence vector \eqref{eq:influe} we can define the
reverse influence vector $I^R$, which we call the defluence vector.   Based on the defluence vector, we can define the metrics 
$\mathcal{D}_1$--$\mathcal{D}_5$ in the same way as $\mathcal{I}_1$--$\mathcal{I}_5$ in \eqref{eq:i1}--\eqref{eq:i5}.

\subsection{Other LON metrics}

The \emph{fractal dimension} of a pattern is an index of spatial complexity and captures the relationship between the level of detail observed in the pattern against the scale of resolution it is measured with. Many real-world complex systems cannot be characterised by a single dimension, however \cite{mandelbrot1997multifractal}; \emph{multifractal} analysis produces a spectrum of dimensions to describe the object. A group of multifractal dimension metrics were computed in previous literature \cite{thomson2022fractal} on the set of LONs in use for the present study; we therefore include them for comparison with the Laplacian measurements. For space reasons we do not describe the algorithm used to obtain the dimensions here (and indeed it is not the purpose of this study); a full description and all parameters associated with the fractal analysis are available in the original work \cite{thomson2022fractal}.

\section{Experimental Setup}
\subsection{LON construction} \label{sec:LONconstr}
We use a set of LONs from the literature \cite{thomson2022fractal} which are extracted from 40 QAPLIB instances of sizes between 25-50. These were constructed using multiple iterated local search (ILS) runs; this is St{\"u}tzle's iterated local search written in C, and is used for both gathering performance data and as the foundation of LON construction \cite{stutzle2006iterated}. Random pairwise swaps in the permutation are used as the neighbourhood. For the hill climbing component, a first-improvement pivot rule was deployed; the perturbation operation applies $k$ random swaps. The dataset considered two options for $k$: $\frac{n}{8}$ (low perturbation) and \textbf{$\frac{3n}{4}$} (high perturbation), where $n$ is problem dimension --- thereby producing two sets of LONs (one for each of the two perturbation settings). There are 100 independent ILS runs per perturbation strength and instance, and runs terminate after 10,000 iterations without an improvement to fitness.

\subsection{Algorithm runs} \label{sec:alg_run}
We require algorithm performance data in order to study the relation with the Laplacian metrics. For this, data from the aforementioned study are used, such that a direct comparison can be made between the metrics proposed here and previous work. Two metaheuristics are considered: ILS and robust taboo search (TS). For both, the mutation operation is a random pairwise exchange in the permutation. The ILS was executed in the same configuration as just described for LON construction, except that the termination condition is met when either the best-known fitness is found or 10,000 iterations have passed without improvement. The TS is Taillard's implementation in C of his robust taboo search (ROTS) algorithm for the QAP \cite{taillard1991robust}\footnote{\url{http://mistic.heig-vd.ch/taillard/codes.dir/tabou_qap2.c}}; it uses a best-improvement pivot rule, and the tabu duration is \(8n\); aspiration is set at \(5n^2\); runs terminate when the best-known fitness is found or after 100,000 iterations. ILS and TS are each executed 100 times per instance, and the performance metric is the \emph{performance gap}, defined as the mean obtained fitness as a proportion of the best-known fitness. 

\subsection{Feature selection}
All feature selection and modelling is coded in \textsf{R}. We perform backwards recursive feature elimination (RFE) to obtain feature sets for the predictive models. The learning algorithm is random forest regression, with the performance of iterated local search or tabu search as response variable. The RFE performs 1000 bootstrapping iterations and compares models based on the root mean squared error (RMSE). There are 22 candidate predictors; these comprise the set of seven from a previous study \cite{thomson2022fractal}, pagerank measurements from the literature \cite{herrmann2018pagerank}, SCC features, and the Laplacian metrics proposed in the present work:

\begin{itemize}
\label{enum:metrics}
\item Number of local  and  global optima
\item Search flow towards global optima
(normalised, aggregated incoming weight from edges directed towards global optima)
\item Median fractal dimension  and variance of fractal dimension (proxy for \emph{multifractality})
\item Maximum fractal dimension
 and range of the multifractal spectrum (proxy for \emph{multifractality})
\item Two pagerank centrality metrics as described in previous literature \cite{herrmann2015predicting,herrmann2018pagerank}, \emph{p1}: pagerank for the global optimum, and \emph{p2}: fitness-weighted average pagerank
\item Three SCC metrics: the number of SCCs, the number of nodes in a SCC, and the node-to-SCC ratio
\item Ten metrics related to LD as introduced in Section \ref{laplacian_metrics_description}; these are labelled \(\mathcal{I}_1\)-\(\mathcal{I}_5\) (influence-based) and \(\mathcal{D}_1\)-\(\mathcal{D}_5\) (defluence-based)
\end{itemize}
Because of the limited number of instances in the QAPLIB of moderate size, our number of observations is small: 40. We therefore restrict the number of selected features to a maximum of four: the \emph{one-in-ten} rule \cite{harrell1984regression} for the ratio between features and observations guided us in this choice. 

\subsection{Modelling}
Following feature selection we build the final models with the chosen predictors. To account for the possible effects of a small dataset size (high variance depending on the data split), models are bootstrapped for 1000 iterations with an 80-20 training-validation set split. The learning algorithm is random forest in its default configuration for \textsf{R}; this constitutes 500 trees, with \(\frac{1}{3}N_f\) features included per split (where $N_f$ is the number of features). The forest allows sampling \emph{with} replacement, and the sample size is the number of observations. To summarise model performance, we consider the pseudo \(R^2\): \(1-\frac{MSE}{variance(y)}\), where $y$ is the response variable (algorithm performance). The pseudo \(R^2\) is usually between 0 and 1, and can be interpreted as the proportion of variance explained; it is also possible to have values below zero --- this happens when the model is explains no variance and is useless for prediction. Whenever there is a value below zero, we convert it to zero to preserve the metric's meaning as proportion of variance explained. We also report the RMSE. For both metrics (pseudo \(R^2\) and RMSE), the bootstrap mean and standard error are provided: see Sec.~\ref{sec:Laplace_predict} and Table~\ref{tab:regression-models}. 

\section{Experimental Results}

\subsection{Visualising information flow on LONs}




\begin{figure*}[ht!]
\centering
\includegraphics[trim = 120mm 80mm 84mm 55mm,clip, width=5.8cm, height=4.4cm]{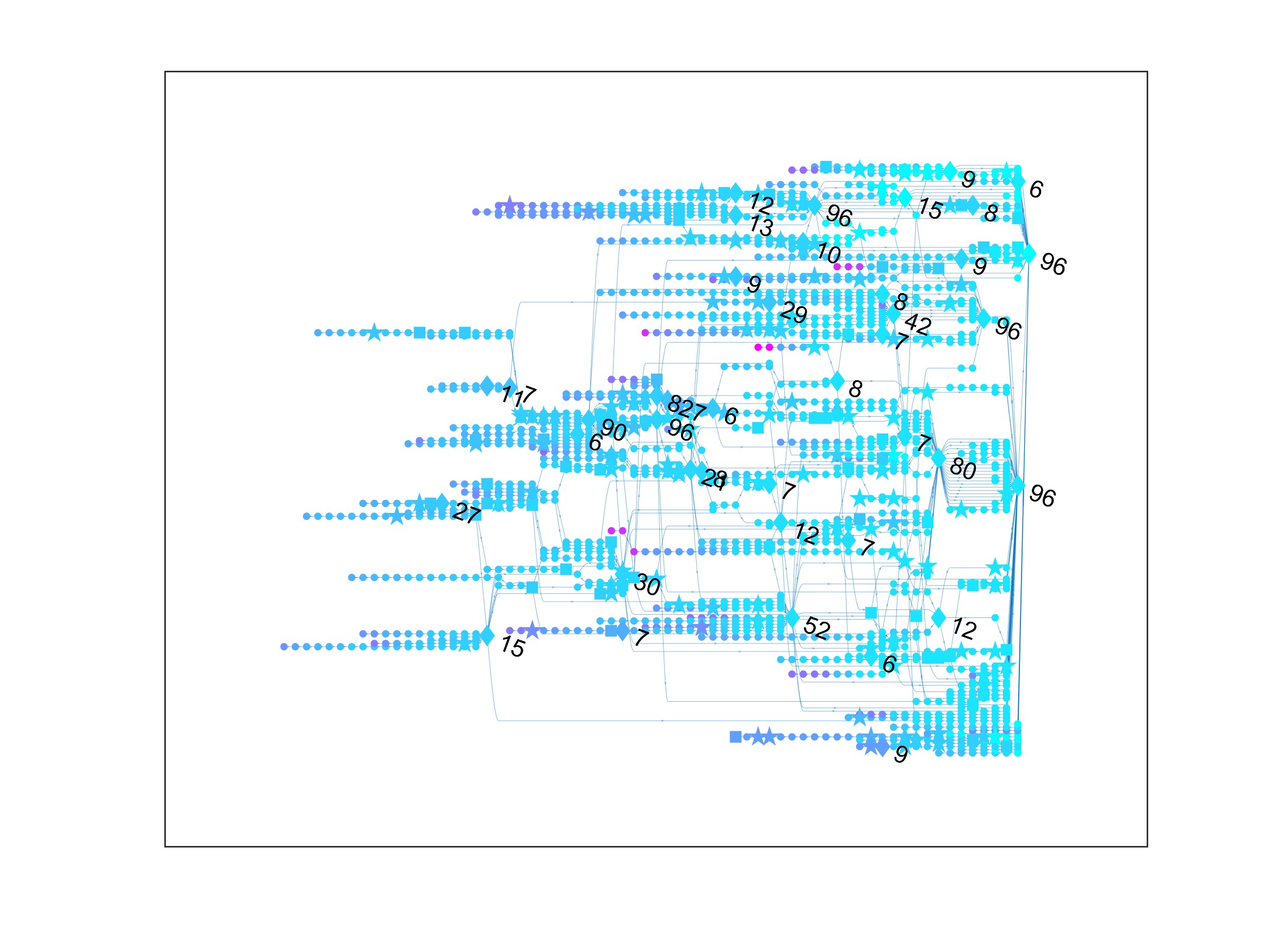}
\includegraphics[trim = 90mm 60mm 64mm 45mm,clip, width=5.8cm, height=4.4cm]{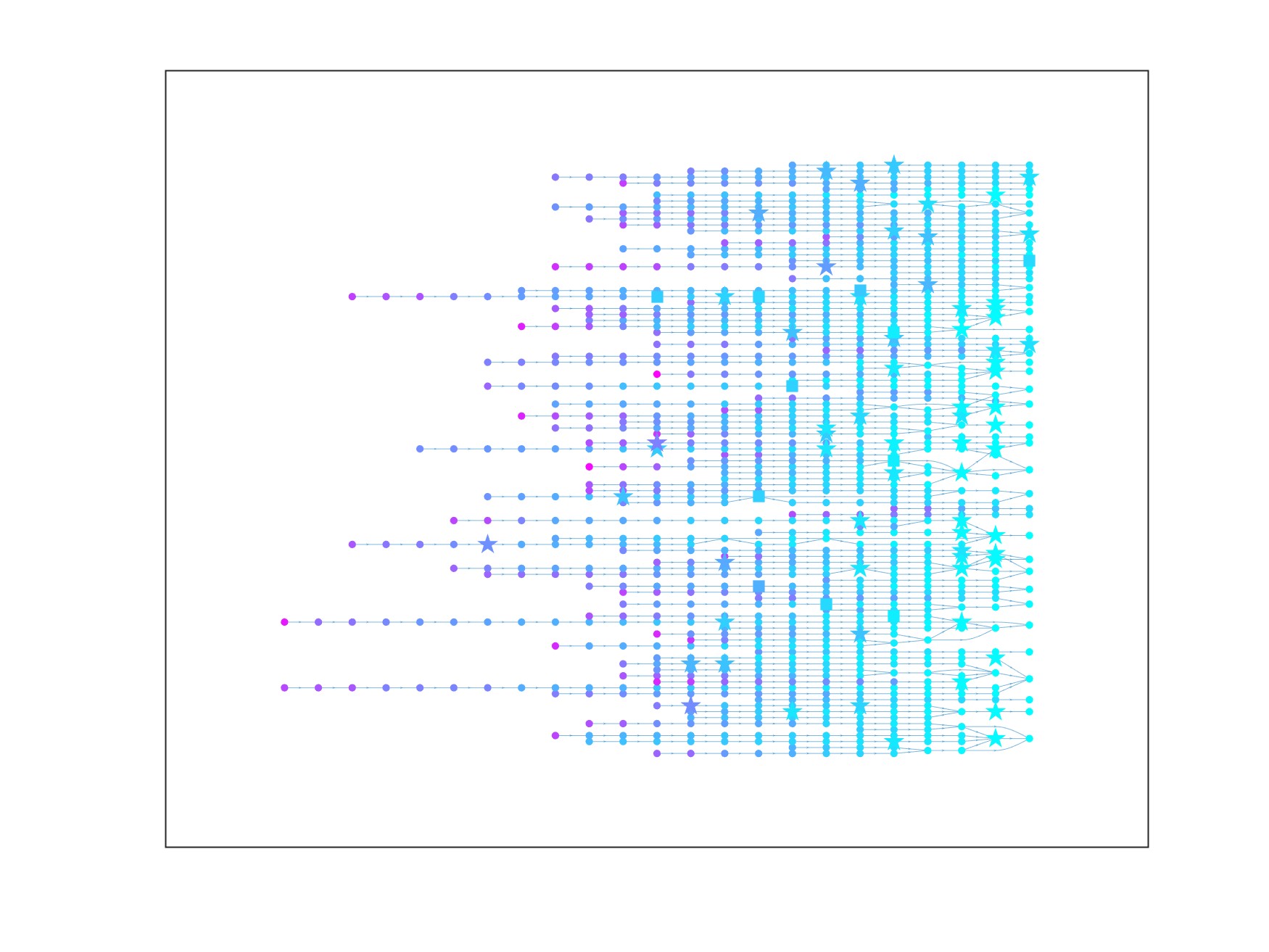}
\includegraphics[trim = 90mm 60mm 55mm 45mm,clip, width=5.8cm, height=4.4cm]{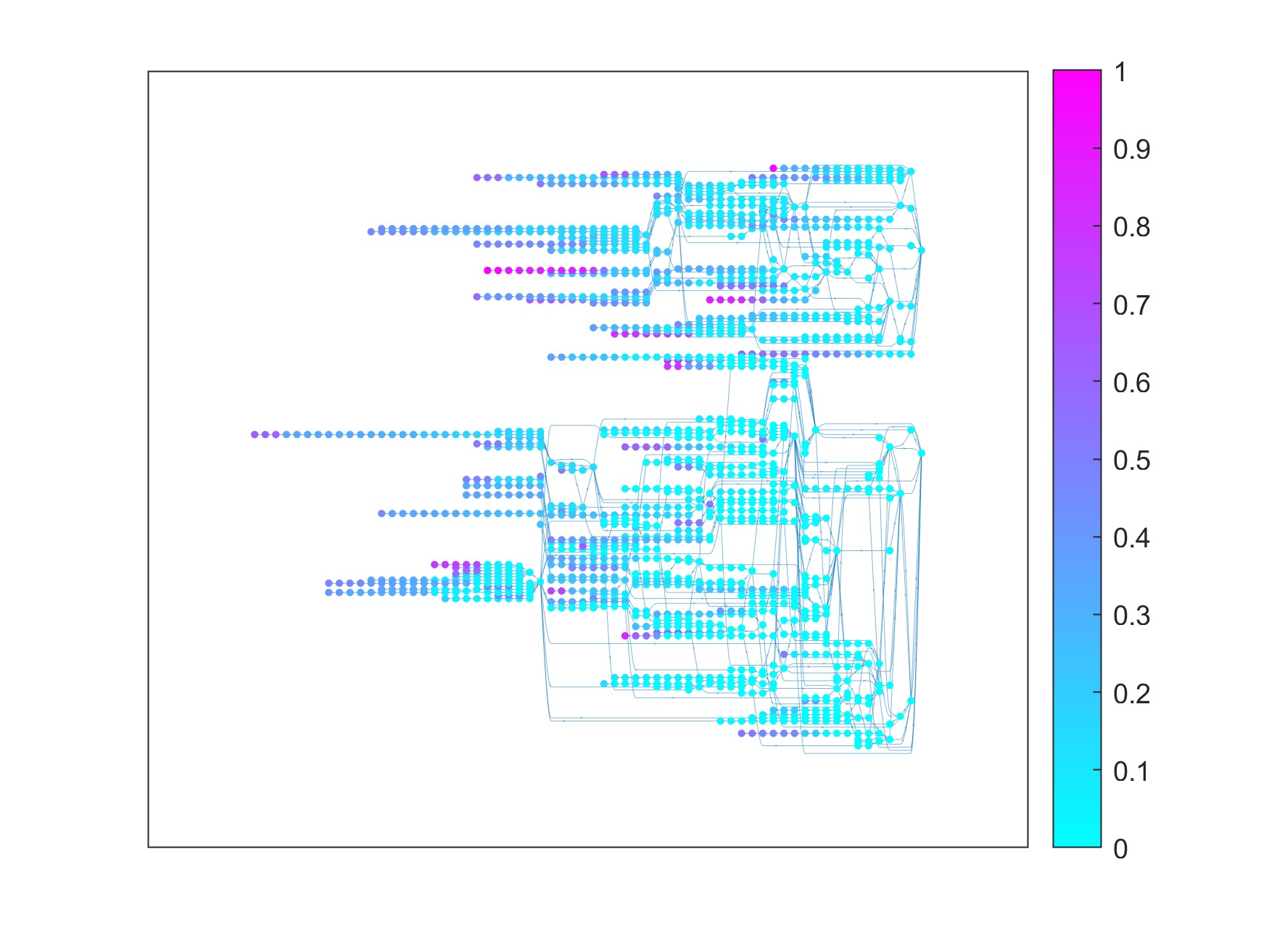} \\
\small Low perturbation \hspace{0.1cm}  (a) bur26a  \hspace{3.75cm} (b) nug25  \hspace{3.95cm} (c) tai30b

\includegraphics[trim = 120mm 80mm 84mm 55mm,clip, width=5.8cm, height=4.4cm]{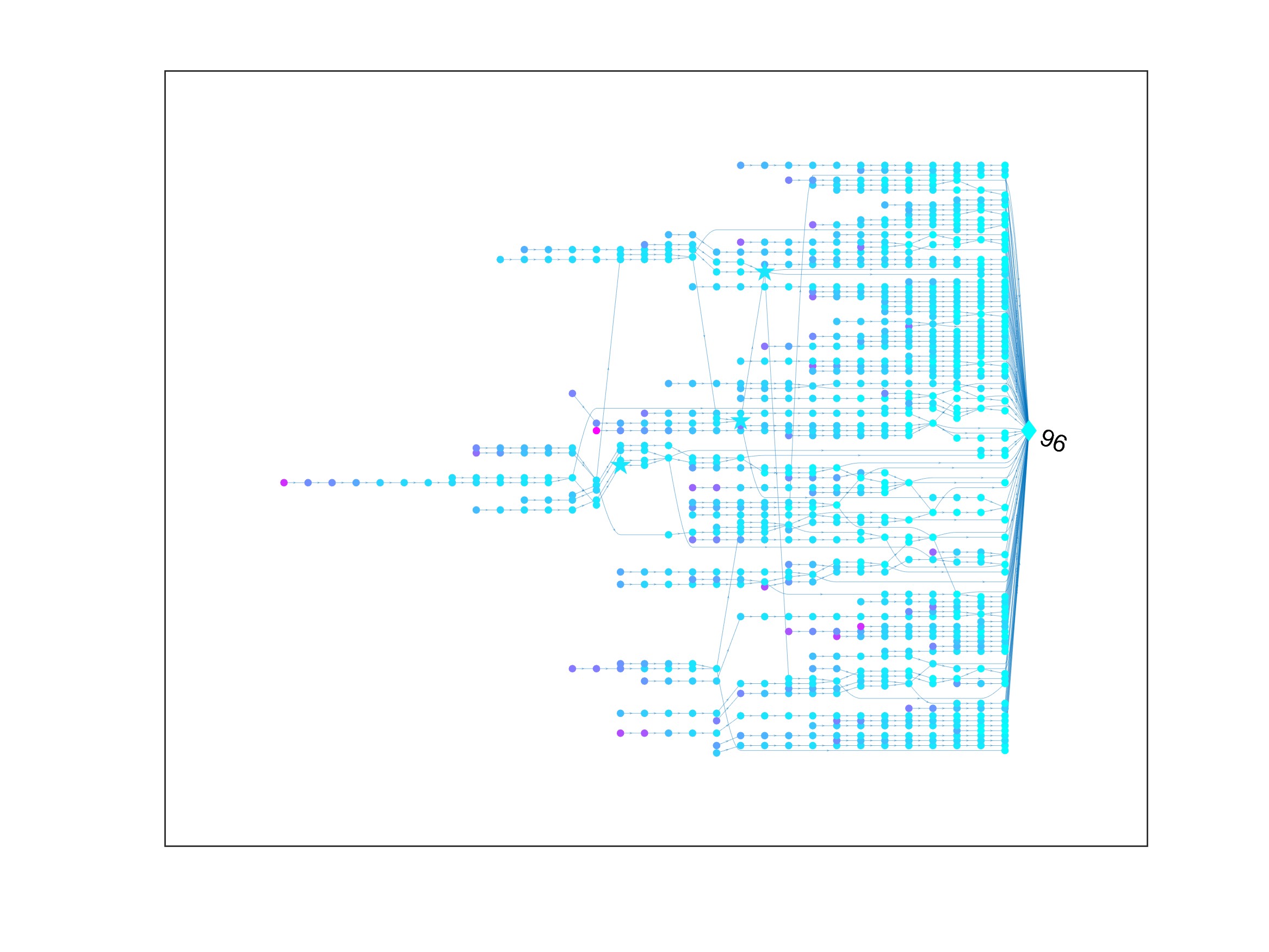}
\includegraphics[trim = 90mm 60mm 64mm 45mm,clip, width=5.8cm, height=4.4cm]{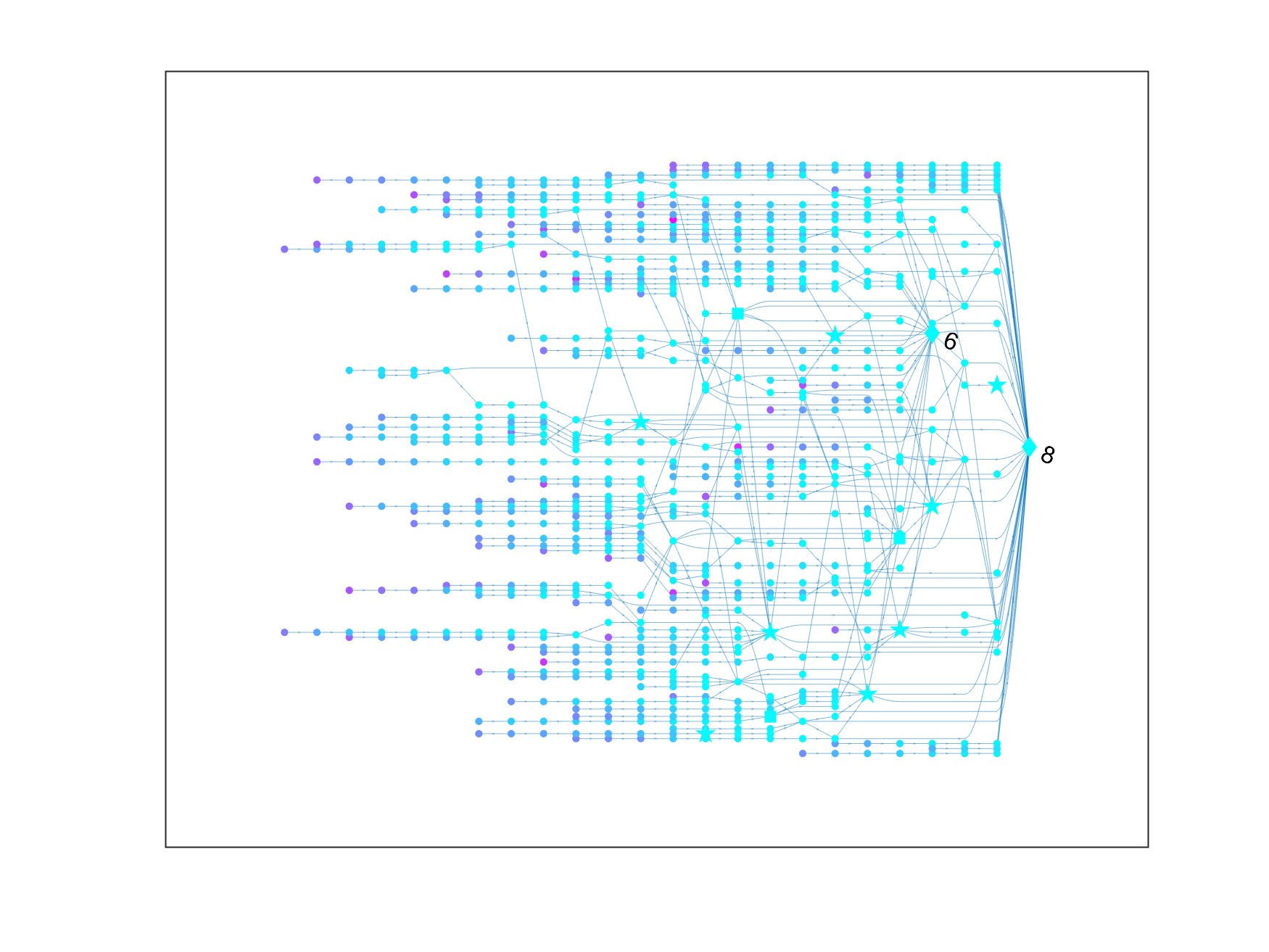}
\includegraphics[trim = 90mm 60mm 55mm 45mm,clip, width=5.8cm, height=4.4cm]{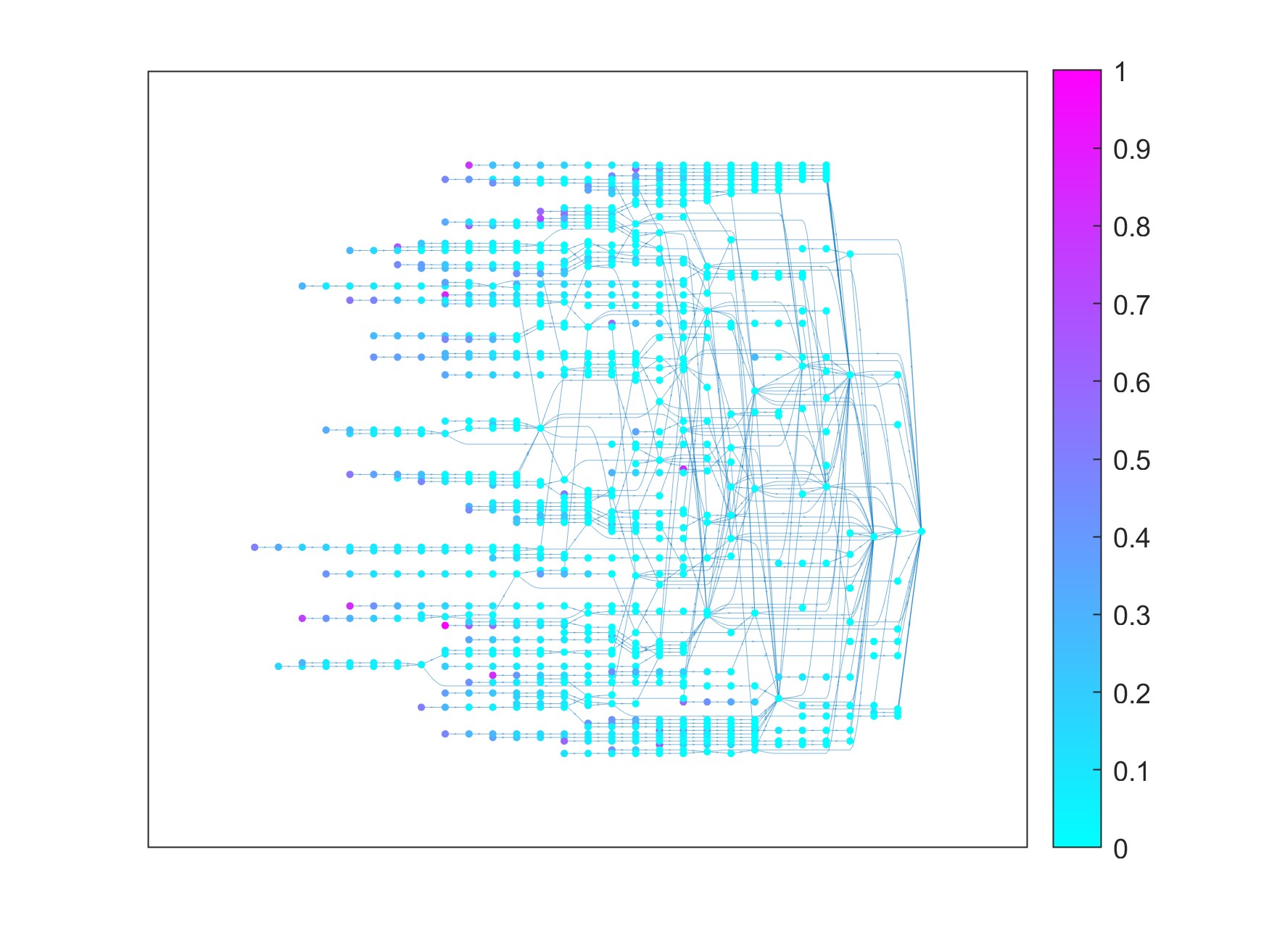} \\
\small High perturbation \hspace{0.1cm}  (d) bur26a \hspace{3.75cm} (e) nug25 \hspace{3.95cm} (f) tai30b

\caption{Typical examples of LONs from QAP instances with both low and high perturbation }
\label{fig:LON_graph_vis_j}

\end{figure*}

Visualization of LONs is a powerful method to gain understanding of underlying network properties. As we here focus on considering LONs as objects describing dynamical processes, we use a visualization accordingly. Fig. \ref{fig:LON_graph_vis_j} gives LONs of three representative QAP instances: a real-world instance {\bf bur26a}, a random flow on grids {\bf nug25} and a random real world {\bf tai30b},  see also \cite{thomson2022fractal}. The LONs in the upper row (Fig. \ref{fig:LON_graph_vis_j}a-c) are for low perturbation, while those in the lower row (Fig. \ref{fig:LON_graph_vis_j}d-f) are for high perturbation. The LONs are shown as to visualize the information flow from source nodes (on the left hand side of the graph) to sink nodes (on the right hand side of the graph). Apart from the nodes shown as dots,  strictly connected components (SCCs) are depicted as stars (for 2 nodes in the SCC), squares (3 nodes) and diamonds (more than 3 nodes). If there are more than 3 nodes in the SCC, we also give the number. The nodes and SCCs are colored according to normalized fitness (with 0 denoting minimal fitness and 1 denoting maximal fitness), with the color code given by the colorbar on the left of the figure. The 6 examples represent typical structural and dynamical  properties of the LONs considered in this paper. All LONs are characterized by rather long paths transferring information from sources to sinks. We see that along the pathway fitness is increasing, sometimes in big leaps. Some of the LONs (bur26a with both low and high perturbation and nug25 with high perturbation) have sinks which form a SCC where all nodes have the same fitness. For bur26a (low and high) the sink SCC has 96 nodes, for nug25 (high) it has 8 nodes. With the exception of tai30b, there are intermediate SCCs between sources and sinks; for bur26a with low perturbation, the number of SCCs and the number of nodes within SCCs is significantly high than for the other examples. Particularly for nug25 (low), the LON has a low degree of network structure and diversity of information flow pathways. Mostly, there are separated paths connecting sources and sinks with few (or no) bifurcations.

\subsection{Properties of SCCs in LONs}
\begin{table*}[ht!]
\centering
\caption{Properties of SCCs in LONs from QAP}
\vspace{2mm}
\footnotesize{
\begin{tabular}{|c|c|c|c|c|c|c|c|} \hline
LON  & Fraction   & Max & Mean  & Max  & Mean  &Max  & Mean   \\ &of LONs& number of& number of&number of&number of&node-to-SCC&node-to-SCC\\ perturbation & with SCC & SCCs& SCCs&nodes in SCC&nodes in SCC&ratio&ratio \\\hline
low & 0.725 & 439 & 181 & 15106 &2466&0.840 & 0.348\\
high &  0.550 & 35 & 6 & 1536&339&0.695 & 0.177\\ \hline

\end{tabular} }
\label{tab:SCC_statistics}

\end{table*}
With Table \ref{tab:SCC_statistics} we give data used for analysing the occurrence of SCCs in the LONs considered in this paper. We give the fraction of LONs (out of 40 considered for low and high perturbation) which possess SCCs, the maximum and mean number of SCCs found in the LONs, the maximum and mean number of nodes in a SCC, and the maximum and mean ratio between the number of nodes in a SCC and the number of nodes in the LON. We see clearly that low and high perturbation strength leads to significantly different values of the considered data.   Thus, also by evaluating these data, a differentiation between low and high perturbation is easily  feasible.  

\subsection{Correlations between performance data and LON metrics}
We next analyse correlations between performance data obtained by ILS and TS, see Sec. \ref{sec:alg_run}, and the Laplacian metrics proposed in this paper as well as LON metrics previously reported, namely median fractal dimension~\cite{thomson2022fractal} and pagerank centrality~\cite{herrmann2015predicting,herrmann2018pagerank}.
\begin{figure*}[ht]
\centering
\includegraphics[trim = 35mm 90mm 43mm 99mm,clip, width=5.8cm, height=3.75cm]{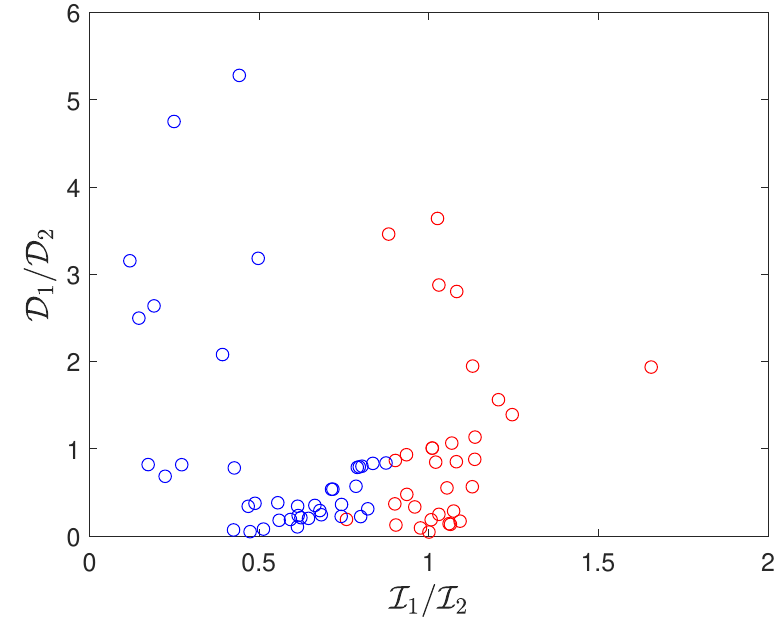}
\includegraphics[trim = 35mm 90mm 43mm 95mm,clip, width=5.8cm, height=3.85cm]{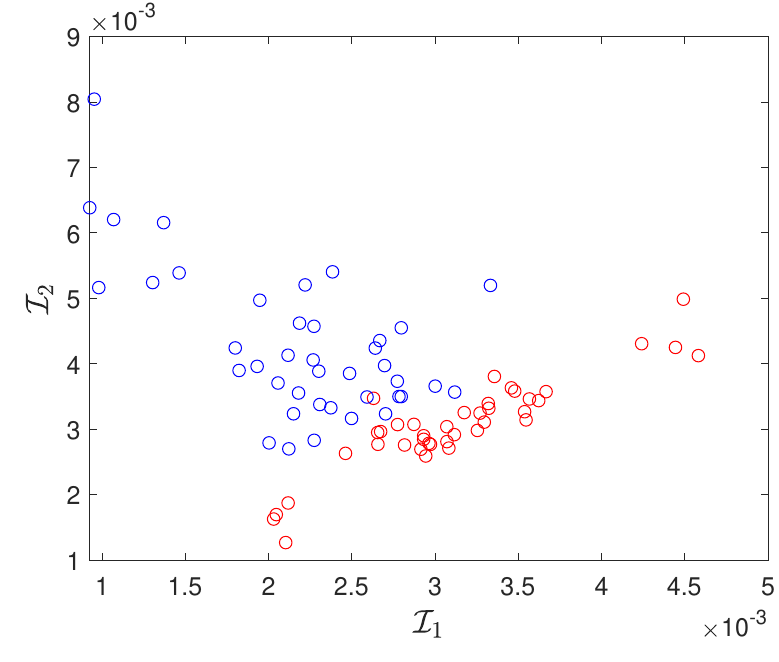}
\includegraphics[trim = 35mm 90mm 43mm 99mm,clip, width=5.8cm, height=3.75cm]{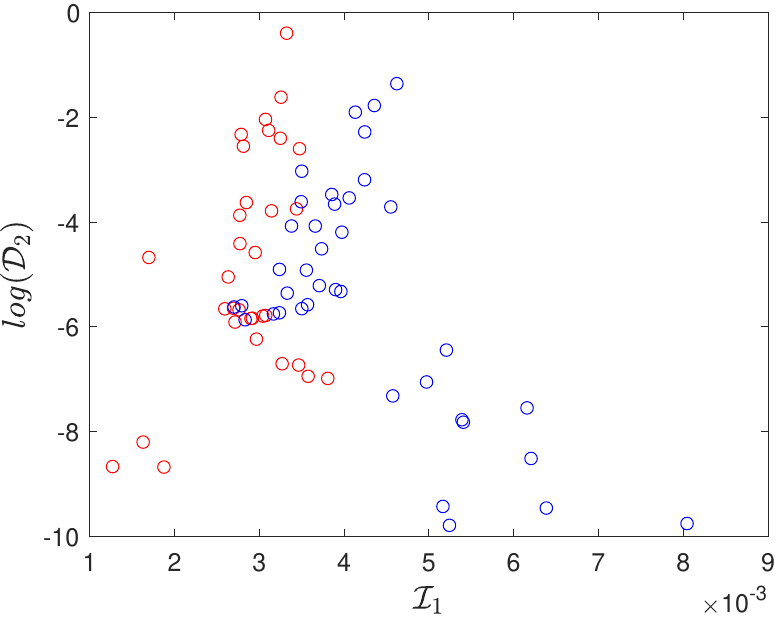}  

\small (a) $\mathcal{I}_1/\mathcal{I}_2$ vs. $\mathcal{D}_1/\mathcal{D}_2$ \hspace{3cm} (b) $\mathcal{I}_1$ vs. $\mathcal{I}_2$ \hspace{3cm} (c) $\mathcal{I}_1$ vs. $log(\mathcal{D}_2)$ 
\caption{Laplacian measures for LONs with different perturbation strength. Clearly differentiable clouds of data enable to classify perturbation strength.  Blue circles indicate low perturbation, red circles high perturbation.}
\label{fig:lapl_measure}

\end{figure*}
We start with relations between the Laplacian metrics proposed in Sec. \ref{laplacian_metrics_description}. Fig. \ref{fig:lapl_measure} shows scatter plots of some of the metrics (or combinations of metrics) for all 40 QAPLIB instances. Low perturbation is given as blue circles, high perturbation as red circles. We see that for the given combinations (but also for other combinations not shown in figures), we get clearly differentiable clouds of data which enables rather straightforwardly to classify low and high perturbation by inspecting the proposed Laplacian metrics.   

\begin{figure}[ht]
\centering
\includegraphics[trim = 35mm 90mm 43mm 99mm,clip, width=8.5cm, height=5.55cm]{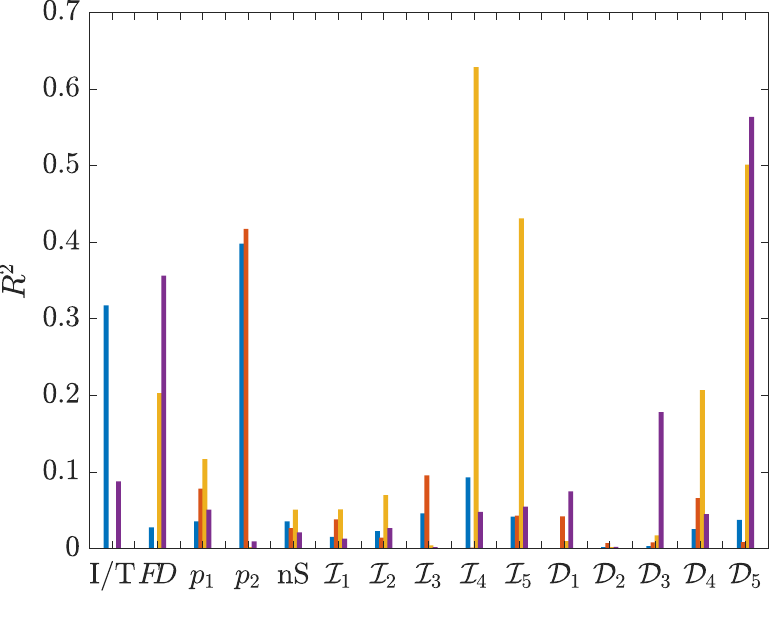}

\small (a) Pearson correlation

\includegraphics[trim = 35mm 90mm 43mm 99mm,clip, width=8.5cm, height=5.55cm]{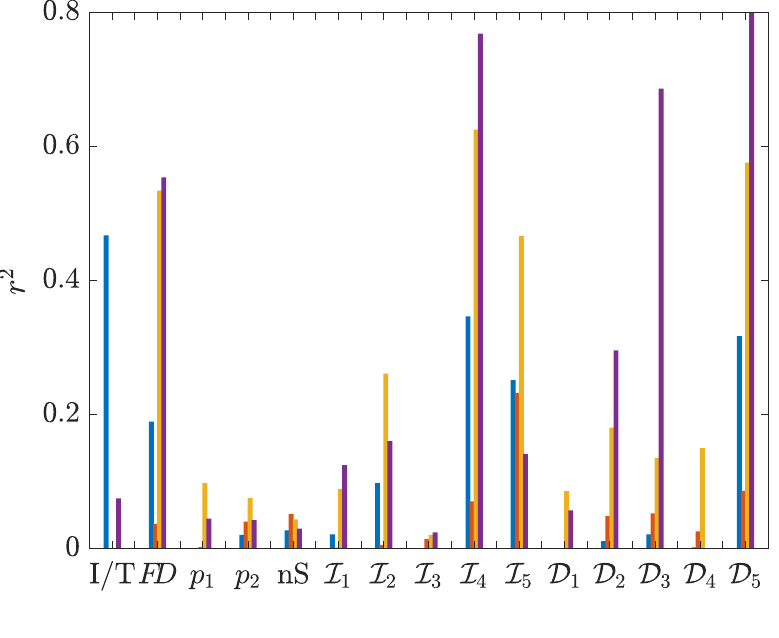}  

\small (b) Spearman rank correlation

\caption{Correlations between TS and ILS performance and LON metrics for both low and high perturbation. $I/T$ is the correlation between ILS and TS performance; median fractal dimension: FD, pagerank: $p_1$, $p_2$, number of SCCs: nS, Laplacian metrics: \(\mathcal{I}_1\)-\(\mathcal{I}_5\) and \(\mathcal{D}_1\)-\(\mathcal{D}_5\). Color code: red-TS (high), blue-TS (low), yellow-ILS (high), purple-ILS (low).}
\label{fig:corr}

\end{figure}

In the following we give correlations between performance data for TS and ILS, and LON metrics for both low and high perturbation, see Fig. \ref{fig:corr}. We test for both linear relationships (Pearson correlation, Fig. \ref{fig:corr}a) and monotonic relationships (Spearman rank correlation, Fig. \ref{fig:corr}b). The colour code for interpreting the plot is available in the Figure caption. Correlations between ILS performance and TS performance, shown as $I/T$ at the far left of the figure serve as a baseline for evaluating the correlations. We notice that some metrics yield high correlations to performance data, even higher than the correlation between ILS and TS performance itself.  Particularly for monotonic relationships between performance and metrics, we get good agreements, for instance for $\mathcal{I}_4$, $\mathcal{I}_5$, $\mathcal{D}_3$ and $\mathcal{D}_5$. Generally, correlation to ILS performance is better than to TS performance with the notable exception of $p_2$ for Pearson and and $\mathcal{D}_4$ for Spearman.

\begin{table*}[ht]
\centering
\caption{Information about models with features selected by recursive feature elimination in a Random Forest setting.}
\vspace{2mm}
\resizebox{0.85\textwidth}{!}{\begin{tabular}{l|cccc}
\toprule
& \multicolumn{2}{c}{Iterated Local Search} & \multicolumn{2}{c}{Robust Taboo Search} \\
\midrule
LON perturbation & low & high & low & high \\ [0.2cm]
\midrule
selected features  &  \begin{tabular}{c}[\(\mathcal{I}_5\), \(\mathcal{D}_4\), \emph{p2}] \end{tabular} &\begin{tabular}{c} [\(\mathcal{I}_4\), \(\mathcal{D}_5\), \emph{Var FD},  \emph{Max FD}] \end{tabular}&  \begin{tabular}{c}[\emph{p2}, \(\mathcal{I}_5\), \(\mathcal{D}_4\), \emph{no scc}] \end{tabular} & [\emph{p2}, \(\mathcal{I}_5\), \(\mathcal{D}_4\), \emph{no scc}]  \\ [0.2cm]
\midrule
\(R^2\)-\emph{train} (SE) & 0.6606 (0.2742) & 0.7725 (0.1947) & 0.7616 (0.2841) & 0.0846 (0.0299) \\[0.2cm]
RMSE-\emph{train} (SE) & 0.0158 (0.0039) & 0.0045 (0.0013) & 0.0673 (0.0281) &  0.0630 (0.0297) \\[0.2cm]
\(R^2\)-\emph{validation} (SE) & 0.9596 (0.3457) & 0.9878 (0.3109) & 0.7820 (0.3404) & 0.8505 (0.3534)\\[0.2cm]
RMSE-\emph{validation} (SE) & 0.0046 (0.0080) & 0.0008 (0.0019) & 0.0325 (0.0448) & 0.0775 (0.0504)\\[0.2cm]
\bottomrule
\end{tabular}}
\label{tab:regression-models}
\end{table*}

\subsection{Laplacian metrics for performance prediction} \label{sec:Laplace_predict}
Table \ref{tab:regression-models} summarises the algorithm performance prediction models. There are four in total --- one for each combination of response variable (ILS or TS) and LON perturbation strength (low or high). In the Table, the \emph{selected features} row indicates, in order of importance, the predictors which were chosen by RFE and therefore used to build the model. The remaining four rows contain model quality metrics: the bootstrap mean of the pseudo \(R^2\) and RMSE for the training and validation sets (the bootstrap standard error is given in parentheses beside the mean). Notice from the \(R^2\)-validation row that these models have mostly high pseudo \(R^2\), which indicates that a majority of variance in the response variable is explained using the predictors. From the \emph{selected features} row, we note that at least two of the Laplacian metrics are selected for each model. In terms of \(R^2\)-validation, three of these four models out-perform those from the previous study~\cite{thomson2022fractal}, and the remaining model is approximately equal. It is interesting that the addition of the Laplacian, pagerank, and SCC metrics into the predictor pool results in the fractal dimension metrics only being selected for one out of four models. The features \(\mathcal{D}_4\), \(\mathcal{I}_5\), and \emph{p2} appear to be particularly strong: they were selected for three out of four model configurations. The number of SCCs, \emph{no scc}, was selected twice. 

\section{Conclusions}
We have proposed a new perspective on local optima networks (LONs). In the past, LON metrics have typically taken a static perspective. Now, instead, we consider the dynamics encoded in them (i.e., the flow of information) with the introduction of Laplacian dynamics (LD) for LONs. As a testbed, 40 instances from the quadratic assignment problem library are used. We extract and propose several measurements related to LD and compare them to previously-proposed metrics from the literature. The results show that some of the new metrics appear to be strong predictors of search difficulty for iterated local search and tabu search, which improve on previously-proposed LON metrics. 
Finally note that the LD approach proposed in the paper solely relies upon rather simple linear algebra operations, for which highly efficient numerical methods are available. Thus, LD metrics scale well with an increasing number of nodes in a LON. 

\section*{Acknowledgment and Data Availability}
We thank Peter Veerman and Ewan Kummel for valuable discussions, and for sharing code for calculating Laplacian kernels.
The network data are available online\footnote{\url{https://t.ly/52nLr}};
metrics will be uploaded to a Zenodo repository upon acceptance. 

\bibliographystyle{IEEEtran}
{\small
\bibliography{example}}

\end{document}